\documentclass[12pt]{article}
\usepackage{graphicx} 
\usepackage{setspace}
\usepackage[margin=1in]{geometry}
\usepackage{hyperref}
\usepackage{amsmath}
\usepackage{subcaption}
\usepackage{biblatex}
\def\keywordname{{\bfseries \emph{Keywords}}}%
\def\keywords#1{\par\addvspace\medskipamount{\rightskip=0pt plus1cm
\def\and{\ifhmode\unskip\nobreak\fi\ $\cdot$
}\noindent\keywordname\enspace\ignorespaces#1\par}}

\title{Influence of various text embeddings on clustering performance in NLP}
\author{Rohan Saha}
\date{April 2023}

\begin{document}

\onehalfspacing

\maketitle

\begin{abstract}
    With the advent of e-commerce platforms, reviews are crucial for customers to assess the credibility of a product. The star ratings do not always match the review text written by the customer. For example, a three star rating (out of five) may be incongruous with the review text, which may be more suitable for a five star review. A clustering approach can be used to relabel the correct star ratings by grouping the text reviews into individual groups. In this work, we explore the task of choosing different text embeddings to represent these reviews and also explore the impact the embedding choice has on the performance of various classes of clustering algorithms. We use contextual (BERT) and non-contextual (Word2Vec) text embeddings to represent the text and measure their impact of three classes on clustering algorithms - partitioning based (KMeans), single linkage agglomerative hierarchical, and density based (DBSCAN and HDBSCAN), each with various experimental settings. We use the silhouette score, adjusted rand index score, and cluster purity score metrics to evaluate the performance of the algorithms and discuss the impact of different embeddings on the clustering performance. Our results indicate that the type of embedding chosen drastically affects the performance of the algorithm, the performance varies greatly across different types of clustering algorithms, no embedding type is better than the other, and DBSCAN outperforms KMeans and single linkage agglomerative clustering but also labels more data points as outliers. We provide a thorough comparison of the performances of different algorithms and provide numerous ideas to foster further research in the domain of text clustering.

\end{abstract}
\section{Introduction}
\label{sec:introduction}
E-commerce websites are prevalent in the digital era and placing orders online for commodities is seamless. E-commerce websites have a feedback system where the customers can submit a review for a product, which usually includes a star rating along with a text review. These reviews are intended to represent the sentiments of the customer towards that specific product. A higher rating for a product indicates that the product serves its intended purpose, which in turn builds the customer's trust in that product. Moreover, the ratings and the corresponding reviews help business improve their products. However, there are instances where the assigned rating does not reflect the implied sentiment of the review. For example, a review with a rating of two may have a review text that might better suited for a four star rating. In other words, mismatches between the rating and review text's sentiment may mislead a potential customer and possibly be detrimental to the business. Using a clustering algorithm can help us group reviews of similar sentiment and potentially assign a rating appropriate to the underlying sentiment of a review. But how do we select a clustering algorithm for such data? And what type of data representation do we use to characterize the review text? We investigate such questions in this work. \par

In the domain of text mining, clustering algorithms have been widely used to find underlying patterns in data (\cite{Aggarwal2012, Subakti2022, density-based-text-clustering, GallardoGarca2020} to name a few). Such works focus on specific text clustering tasks and use particular types of numerical vectors to represent the text. There is a paucity of work exploring the impact of different text representations' on the performance of different types of clustering algorithms. In other words, how do we choose a text representation for a specific type of clustering algorithm? Our work presents a crucial step in this direction. \par

In this work, we compare the performance of various types of clustering algorithms when applied to different types of text representations (embeddings). We obtain textual data from product reviews from Amazon.com and represent them using different types of embeddings. For each type of embedding, we train four types of clustering algorithms, namely partitioning, hierarchical, and two density based, and compare the performance of each algorithm using internal and external validation techniques. Our results indicate that the choice of text representation has a drastic effect on the performance of a clustering algorithm, and density based algorithms may perform better than other types of clustering algorithms. We also found that it may not always be the case that the number of clusters identified by the clustering algorithms is equal to the number of predefined labels (in our case, the labels are the ratings assigned to each review). All data and code is available at \url{https://github.com/simpleParadox/cmput_697_project}


\section{Methods}
\label{sec:methods}
\begin{figure}[!htb]
    \centering
    \includegraphics[width=1.0\textwidth]{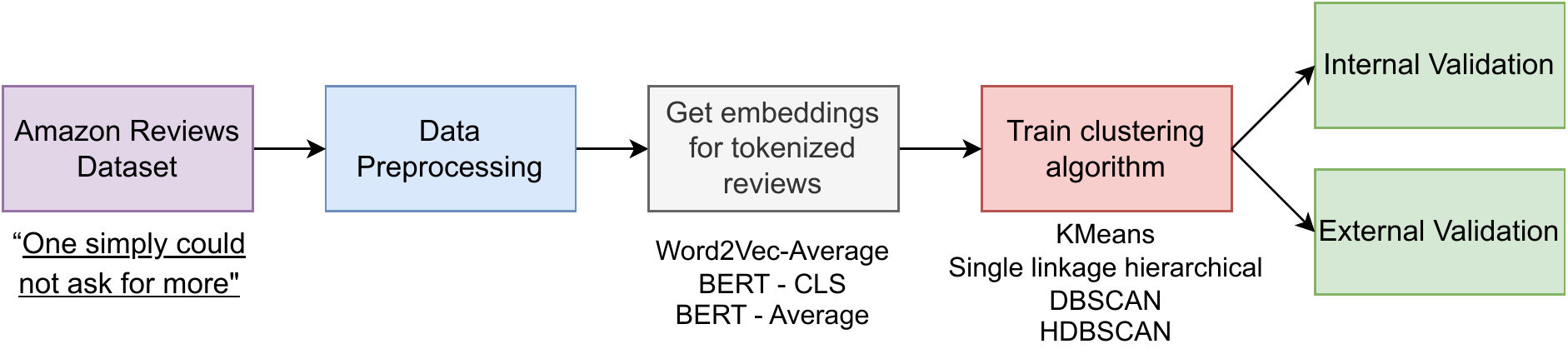}
    \caption{Experimental framework diagram. First, the reviews are loaded and preprocessed, where the tokenization of reviews takes place. Then, various embeddings are obtained for each review on which different clustering algorithms are trained. To evaluate the performance of the clustering algorithm, we use internal validation (silhouette score) and external validation (adjusted rand index score and cluster purity).}
    \label{fig:experimental_framework}
\end{figure}

In this section, we explain the different components of our experimental paradigm. For simplicity we show an overview of the experimental framework in Figure \ref{fig:experimental_framework}.

\subsection{Dataset and Preprocessing}
\label{sec:dataset}
\begin{figure}
    \centering
    \includegraphics[width=1.0\textwidth]{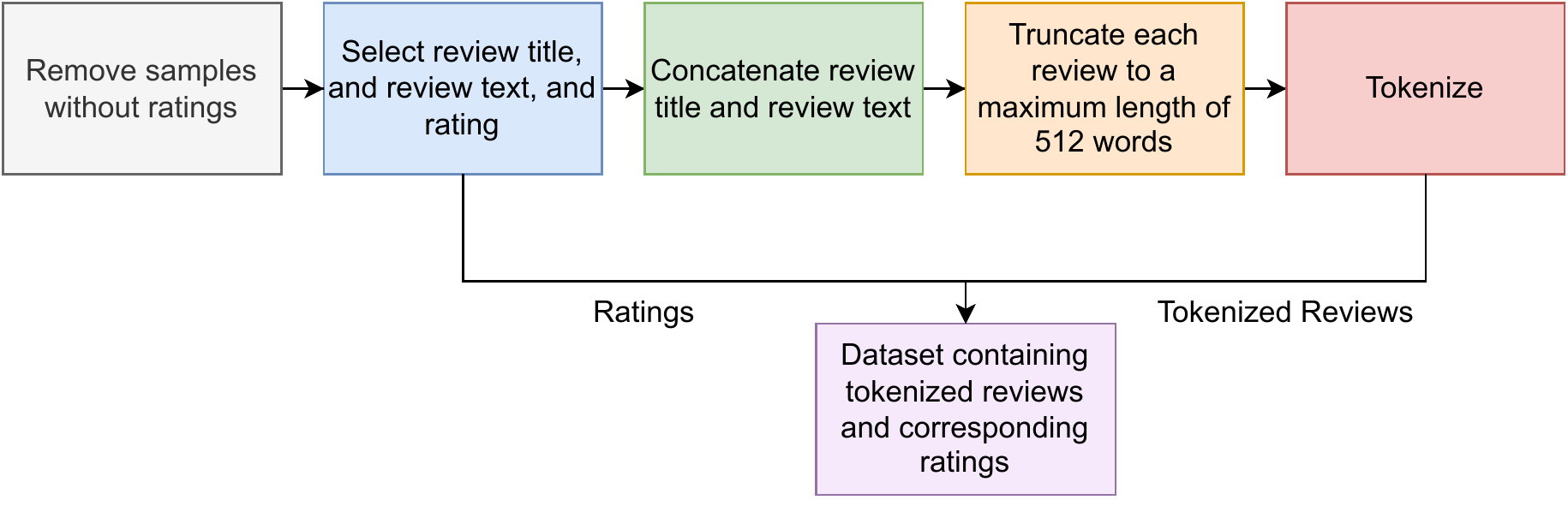}
    \caption{Preprocessing framework depicting the steps involved in data cleaning. First, we remove the reviews without a rating. We then select the review title and review text, and concatenate the two. We truncate the concatenate sample to have a maximum length of 512 words. Finally, we tokenize the input to obtain the text representations. A dataset is then formed that contains the data.}
    \label{fig:preprocessing_framework}
\end{figure}

\begin{figure}[!htb]
    \centering
    \includegraphics[width=0.8\textwidth]{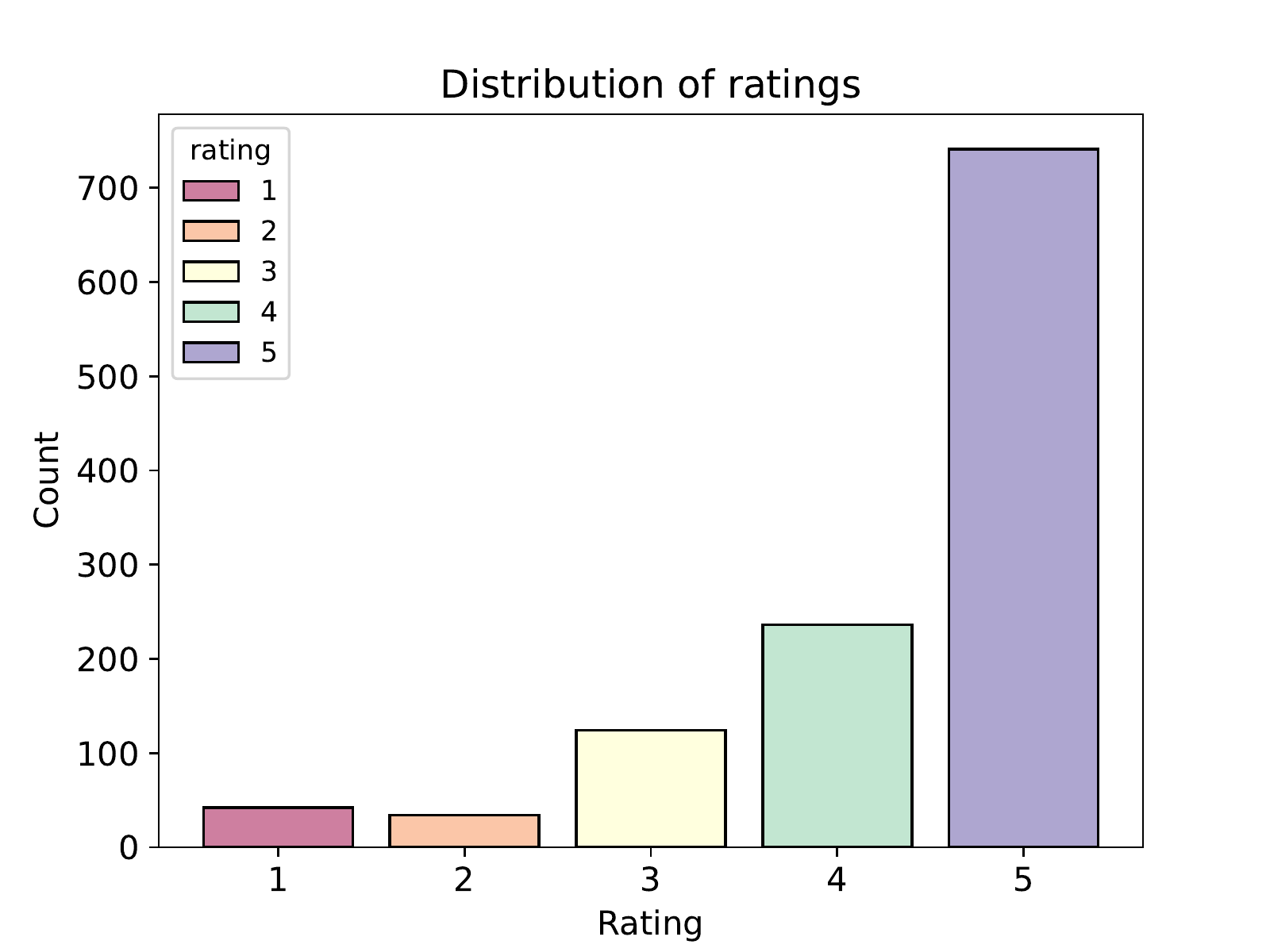}
    \caption{Distribution of ratings in the dataset. The majority of the reviews in the dataset are skewed towards higher ratings.}
    \label{fig:ratings_distribution}
\end{figure}

In this work, we use the Amazon product reviews dataset\footnote{\url{https://www.kaggle.com/datasets/yasserh/amazon-product-reviews-dataset}} that contains 1597 samples and 27 columns of primarily consumer electronics from the Amazon brand. We remove the samples where the rating is not present for a review. The final dataset contains 1177 samples. Each sample contains the product name, product brand, number of stars for the rating, review text, review title, etc. (refer to the dataset link for the full description of each feature). Given the scope of this work, we will only use the review text, review title, and the star rating accompanying each review. The star ratings are in the range of 1-5. We concatenate the review title and the review text to represent the samples in the dataset. The concatenation ensures that the review title contributes to the overall information of the review. We also truncate each concatenated sample to have a maximum 512 tokens as this is required with the maximum length of the input to a language model (discussed in Section \ref{sec:embeddings}). We show the preprocessing steps in Figure \ref{fig:preprocessing_framework}. 
We show the distribution of the ratings in Figure \ref{fig:ratings_distribution}. We observe that the ratings in the dataset are not uniformly distributed.

\subsection{Embeddings}
\label{sec:embeddings}
Our goal is to investigate the effect of different types of text embeddings on the performance of clustering algorithms. To this end, we use two language models the pretrained BERT \cite{devlin2019bert} (pretrained on the BookCorpus dataset and the English Wikipedia dataset) and pretrained Word2Vec \cite{mikolov2013efficient} (pretrained on the Google News dataset) to represent each review in the dataset. We use the Word2Vec implementation from Gensim \cite{rehurek_lrec} and the BERT implementation from Huggingface\footnote{\url{https://huggingface.co/bert-base-uncased}}. For each sample, we concatenate the review title and the review text, before feeding them into the language models. For Word2Vec, we first tokenize the sample and then obtain the hidden vector (embedding) for each token (word) from the language model. Each hidden vector is of size 300 dimensions. Finally, we average the hidden vectors for each token to obtain one single 300 dimensional vector for the review sample.
Word2Vec embeddings do not capture contextual information present in the review. Therefore, we also use BERT to account for contextual variations in the review text as previous work has shown that BERT embeddings outperform other text embeddings \cite{Subakti2022}. To obtain BERT embeddings, we tokenize the sample using the pretrained BERT tokenizer and then feed the tokenized result into the BERT model. We use two types of BERT embeddings to represent each review. First, we use the \texttt{<CLS>} token to represent the review which is a single 768-dimensional vector that takes into consideration all the tokens in the input review text. Second, we obtain the last hidden state (768-dimensional) vector for each token, which we average to obtain a single 768-dimensional vector. For each type of embedding, we use StandardScaler from scikit-learn \cite{sklearn_api} to z-score the data.
To summarize, we use three types of embeddings.
\begin{itemize}
    \item Word2Vec - Average - 300 dimensional.
    \item BERT - CLS - 768 dimensional.
    \item BERT - Average - 768 dimensional.
\end{itemize}
\par
\begin{figure}[!htb]
    \centering
    \hspace*{\fill}
    \hspace*{\fill}
    \hspace*{\fill}
    \begin{subfigure}{0.80\textwidth} 
      \includegraphics[width=0.9\textwidth]{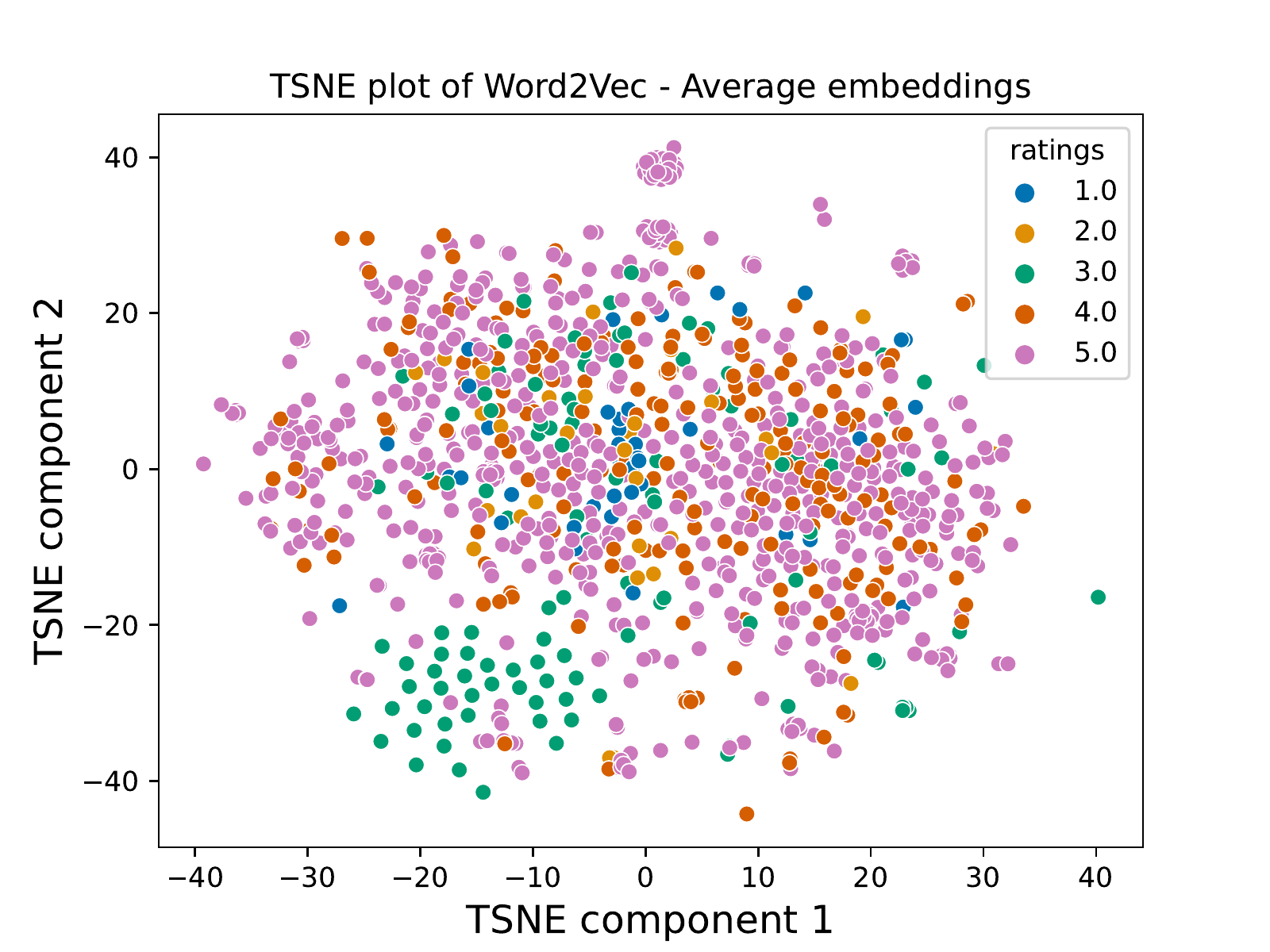}
      \caption{}
      \label{fig:w2v_embeddings_tsne} 
    \end{subfigure}
    \hspace*{\fill} 
    \par\vspace{1em}
    \begin{subfigure}{0.48\textwidth}
      \includegraphics[width=\textwidth]{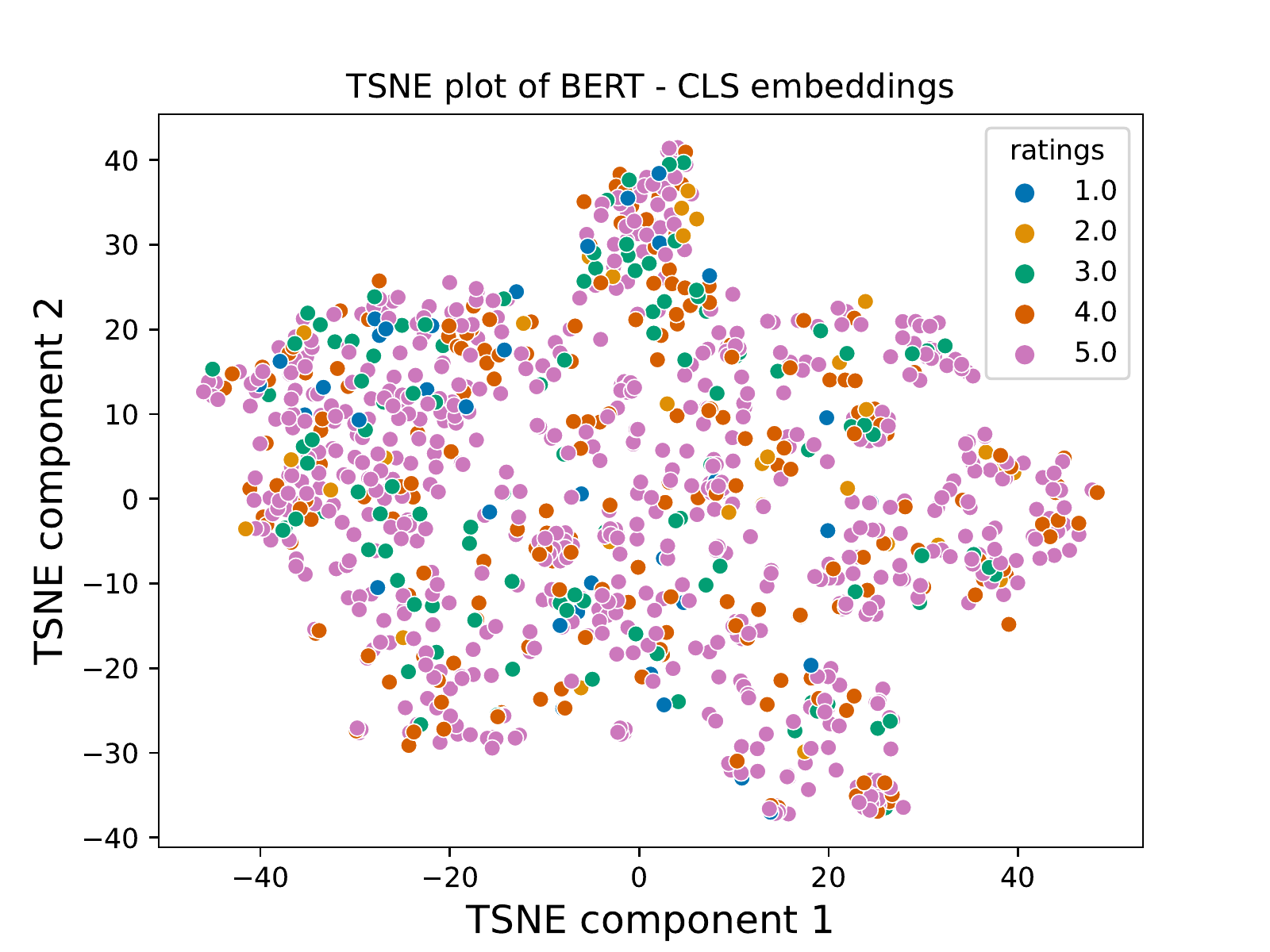}
      \caption{} 
      \label{fig:bert_embeddings_tsne}
    \end{subfigure}
    ~ 
    \begin{subfigure}{0.48\textwidth}
      \includegraphics[width=\textwidth]{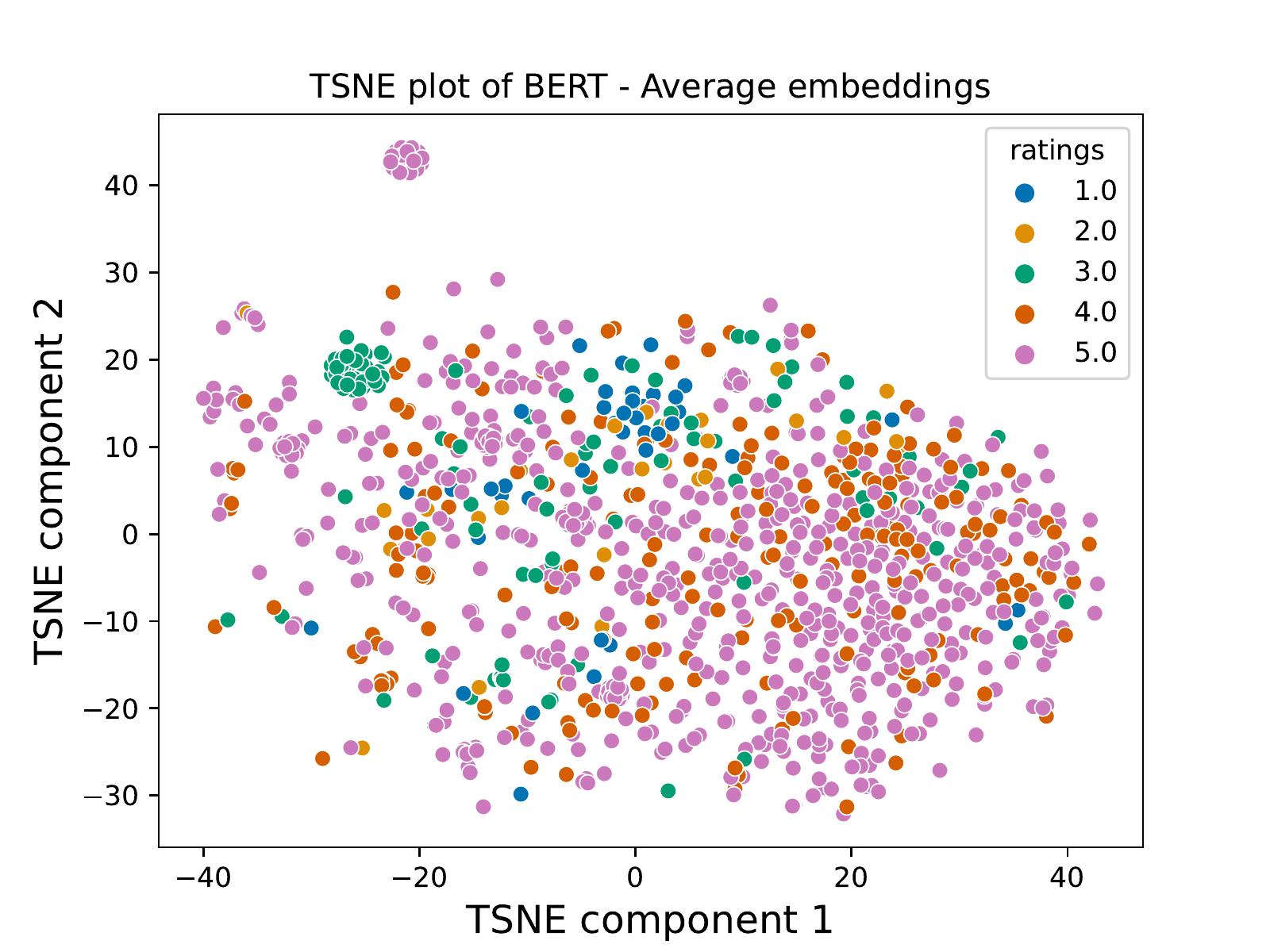}
      \caption{} 
      \label{fig:bert_avg_tsne}
    \end{subfigure}
    \caption{TSNE plots in 2 - dimensions for the different embedding types used in this work. Figure \ref{fig:w2v_embeddings_tsne} shows the distribution of the Word2Vec embeddings, Figure \ref{fig:bert_embeddings_tsne} shows the distribution of the BERT - CLS embeddings, and Figure \ref{fig:bert_avg_tsne} shows the distribution of the BERT - Average embeddings. None of the embeddings are similarly distributed in space.}
    \label{fig:tsne_plots_for_embeddings}
  \end{figure}

Each embedding type encodes the text differently. Moreover, as embeddings vary in the number of dimensions, their distribution in space is also different. To visualize the distribution of the embeddings in space, we use a dimensionality reduction technique to plot the embeddings in 2 - dimensions. We apply t-SNE \cite{JMLR:v9:vandermaaten08a} with 2-components on each embedding type. We show the plots for each embedding in Figure \ref{fig:tsne_plots_for_embeddings}. We observe that each embedding type is distributed differently in space. This characteristic of the data may result in the varied performance of different clustering algorithms.

\subsection{Clustering Algorithms}
\label{sec:clustering_algorithms}
To find patterns in the vector representations of the reviews, we apply four clustering algorithms and compare each algorithm's performance. We use KMeans \cite{k-means_1, k-means_2}, single linkage agglomerative hierarachical \cite{müllner2011modern} as partitioning and hierarchical based methods respectively, and DBSCAN \cite{dbscan}, and HDBSCAN \cite{Campello2013} for density based methods. We use the implementation from scikit-learn \cite{sklearn_api} for KMeans, Agglomerative Hierarchical single linkage clustering, and DBSCAN. For HDBSCAN, we use the implementation from the \texttt{hdbscan} library \cite{mcinnes2017hdbscan}.

\subsection{Evaluation}
\label{sec:evaluation}
Numerous validation metrics have been proposed to evaluate the performance of clustering algorithms (see \cite{wegmann2021review} and \cite{Xu2015} for a review). The validation metrics are generally categorized into internal and external validation. For internal validation, we use the silhouette score\footnote{\url{https://scikit-learn.org/stable/modules/generated/sklearn.metrics.silhouette_score.html}}. For external validation, we use the adjusted rand-index\footnote{\url{https://scikit-learn.org/stable/modules/generated/sklearn.metrics.adjusted_rand_score.html}} and cluster purity, which measures the extent to which a cluster contains a single class.

\subsection{Experimental Settings}
\label{sec:experiments}

Our dataset has text reviews for which the ratings range from a value of 1 to a value of 5. For the KMeans and the single linkage agglomerative hierarchical algorithms we chose to conduct two variations of experiments. First, we set the number of clusters to 5 (one for each rating value) and apply the various clustering algorithms to the dataset. Second, we apply the clustering algorithms on the dataset as if there were 3 clusters. We transform the ratings to have three class labels by applying the following transformation.
\begin{equation*}
    \text{rating} = \begin{cases}
    1 & \text{if rating} < 3 \\
    2 & \text{if rating} = 3 \\
    3 & \text{if rating} > 3
    \end{cases}
\end{equation*}
It is useful to note that we only specify the number of clusters as a hyperparameter for the KMeans and the Agglomerative Hierarchical Clustering.
\par
For KMeans, we initialize the algorithm with the k-means++ method where the initial cluster centroids are selected using a sampling based approach from the empirical probability distribution of the points' contribution to the overall inertia\footnote{\url{https://scikit-learn.org/stable/modules/generated/sklearn.cluster.KMeans.html}}. We set the maximum number of iterations to 300. For DBSCAN, we tune the hyperparameter \texttt{epsilon} $\epsilon$, which controls the maximum distance between two samples for them to be considered in the neighbourhood of each other, and \texttt{min\_samples} is fixed to a value of 5, which defines the number of samples in the neighbourhood of a point for it to be considered a core point. For the HDBSCAN algorithm, we only tune the parameter \texttt{min\_cluster\_size}, which controls the minimum number of points for a cluster to form. We discuss the values for the hyperparameters in the Results section (Section \ref{sec:results}). For all the algorithms, we use euclidean distance metric.

\section{Results and Discussion}
\label{sec:results}
Various types of embeddings may affect the performance of different clustering algorithms, for which we investigate the results in this section.

\subsection{KMeans}
\label{results:kmeans}
We start by exploring the effect of choosing various embedding types on the KMeans algorithm. One of the crucial hyperparameters of the KMeans algorithm is the number of clusters, as KMeans requires the user to specify the number of clusters in the data. The review dataset contains reviews categorized into five categories (ratings). However, this categorization may not reflect itself in terms of the number of clusters in the embedding space. The question then arises, what value should we specify for the number of clusters? One of the popular methods to choose the optimal number of clusters is the elbow (or knee) method \cite{Thorndike1953} where the inertia (within cluster sum of squares) is graphed on the ordinate (y-axis) and the number of clusters on the abscissa (x-axis); the point on the graph where there's a 'break' or 'knee' is chosen as the optimal number of clusters. However, when we plotted the inertia against the number of clusters, there was no discernible break in the graph. Therefore, we chose the silhouette score method to determine the optimal number of clusters.
\par

\begin{figure}[!htb]
    \centering
    \includegraphics[width=0.8\textwidth]{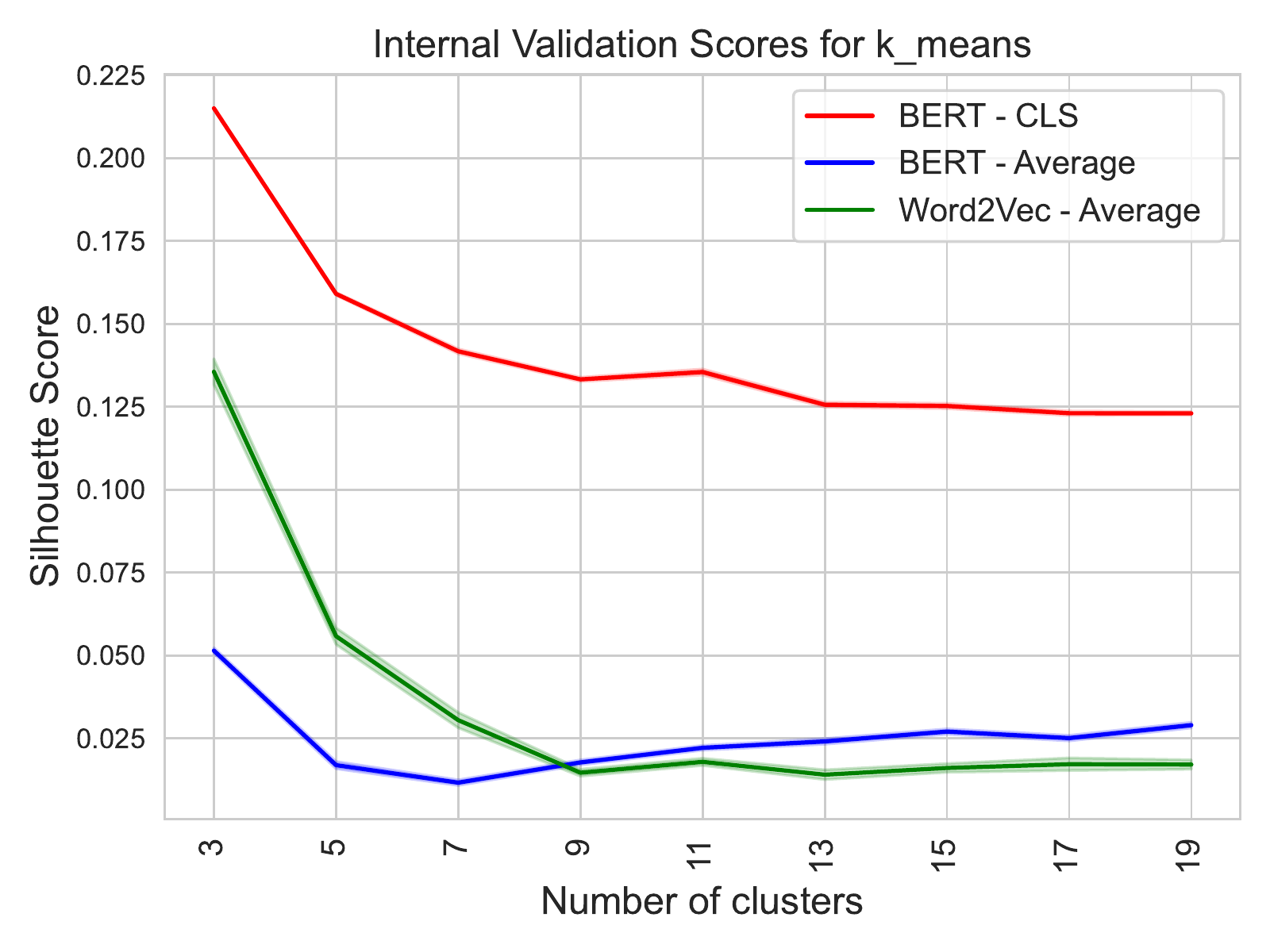}
    \caption{Internal validation scores for KMeans with varying number of clusters.  Since the initialization centroids of the KMeans affects the performance of the algorithm, we run the algorithm with starting seed values in the range of 1-50. The average results are shown here and the standard error of mean is shown with shaded areas along the curves.}
    \label{fig:internal_validation_kmeans}
\end{figure}

The silhouette score is a metric generally used for internal validation of clustering algorithms but it can also be used to choose the number of clusters (\texttt{n\_clusters}) for the KMeans algorithm \cite{schubert2022stop}. We tested a range of values (3 - 19) and plotted the silhouette score for each value. Since the performance of the KMeans algorithm also depends on the initialization of the cluster centroids, for each value of \texttt{n\_clusters}, we used seed values between 1 - 50 to initialize the cluster centroids. In Figure \ref{fig:internal_validation_kmeans}, we show the internal validation scores for KMeans for each embedding type, averaged across multiple seeds, and the standard error of mean shown along the curves. 

\begin{figure}[!htb]
    \centering
    \includegraphics[width=0.8\textwidth]{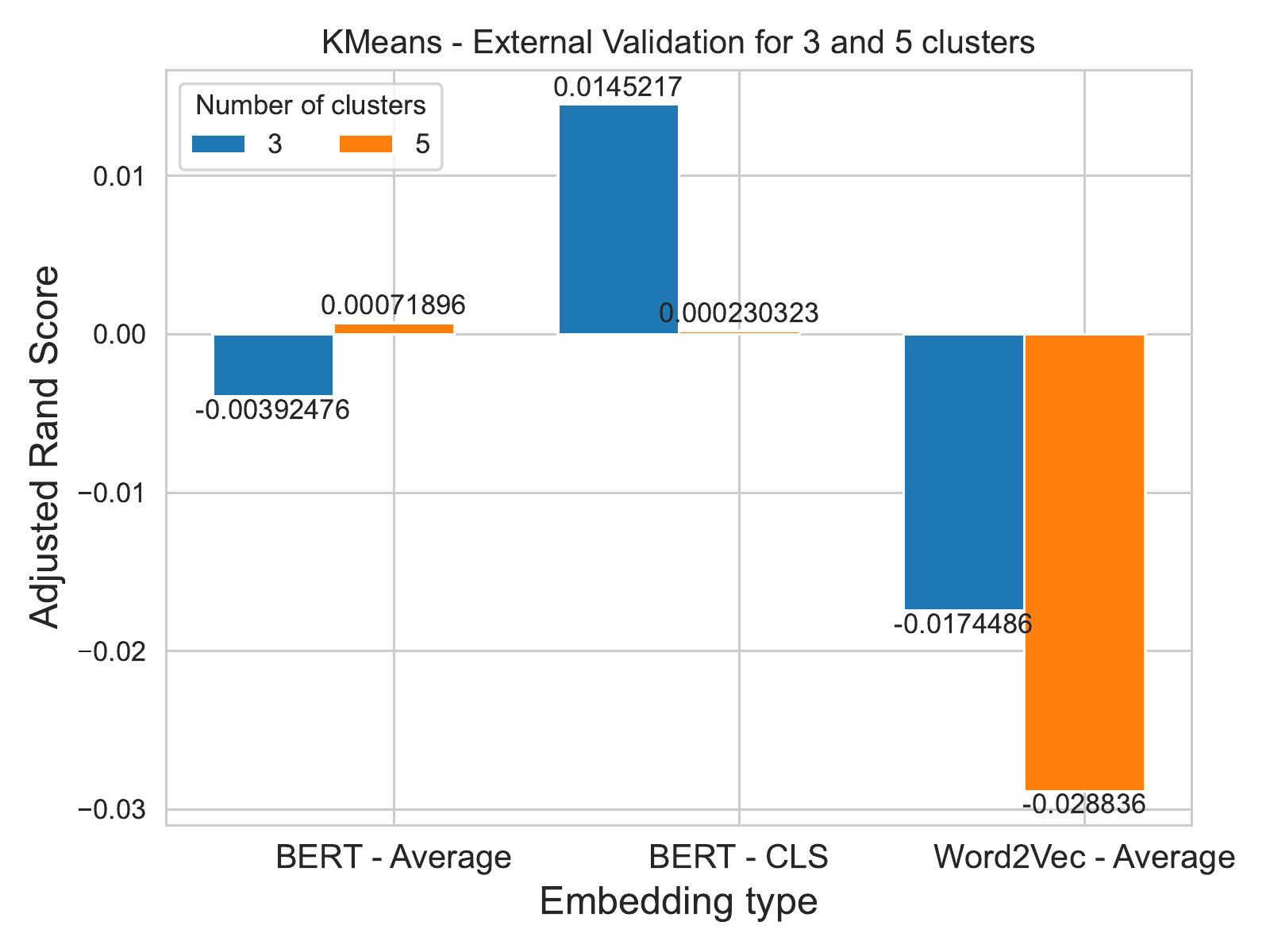}
    \caption{External validation results for KMeans averaged across fifty seed values (seed 1-50). BERT - CLS embeddings for n\_clusters = 3 seem to outperform other embeddings. However, the values are close to zero indicating poor clustering quality.}
    \label{fig:external_validation_kmeans}
\end{figure}

\begin{figure}[!htb]
    \centering
    \includegraphics[width=0.8\textwidth]{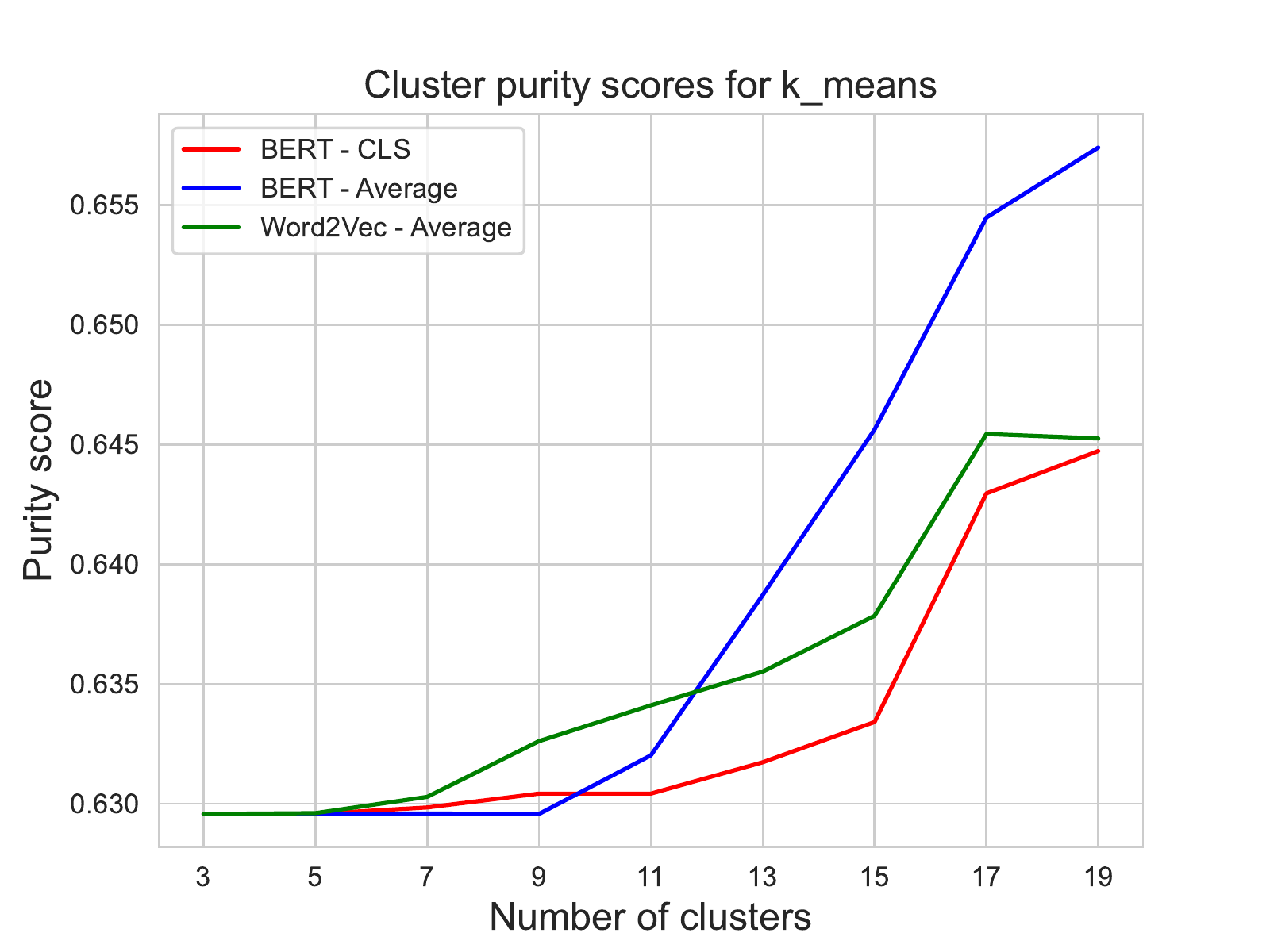}
    \caption{Cluster purity scores for the KMeans algorithm. As the number of clusters increases, the purity scores increases. However, the purity scores of around 0.65 indicate sub-optimal clustering quality.}
    \label{fig:purity_kmeans}
\end{figure}

Each curve on the graph represents an embedding type. The red curve denotes the results when the KMeans algorithm is trained on the BERT - CLS embeddings, the blue curve shows the results for the BERT - Average embeddings, and the blue curve shows the results for the Word2Vec - Average embeddings. We observe that the silhouette score for all three embeddings is maximum when the number of clusters is 3, and this score stabilizes as the number of clusters increases. Such a result indicates that the optimal number of clusters is 3. Our dataset, however, is as five ratings (categories). This brings up the question, does the data belong to three clusters instead of five? To this end, we use external validation to evaluate the performance of the KMeans algorithm for each embedding type.
\par

We show the external validation scores in Figure \ref{fig:external_validation_kmeans} for both three clusters and five clusters. The adjusted rand scores for the BERT - CLS embedding outperform those of BERT - Average and Word2Vec - Average. Although such a result is consistent with the results of the internal validation scores, the difference between the scores is not significant. From the external validation results, it is not clear whether choosing 3 or 5 clusters will result in better model performance. Moreover, the adjusted rand scores for both three and five clusters are close to zero indicating poor clustering quality, thus requiring the need for the evaluation of clustering algorithms. \par

We also use the cluster purity measure as an evaluation method to assess the extent to which a cluster contains a single class. We show the cluster purity results in Figure \ref{fig:purity_kmeans}. The scale of the purity value is important to note. The purity value is the lowest when the number of clusters is close to 3 or 5. As the number of clusters increases, the purity value increases. This is because the more the number of clusters, the fewer points that belong to each cluster. When the number of clusters is equal to 3, the silhouette scores were the highest (see Figure \ref{fig:internal_validation_kmeans}). The purity scores also indicate that the clusters identified by the KMeans algorithm are possibly poor in quality.

All in all, BERT - CLS can be the choice of embedding if KMeans is used to cluster text embeddings, but caution must be exercised as optimal results may not be guaranteed. In the next section, we investigate the performance of agglomerative single linkage clustering algorithm to cluster the text embeddings.


\subsection{Single Linkage Agglomerative Hierarchical}
\label{results:agglomerative}
Although our dataset contains reviews specified into 5 categories (ratings), the KMeans analysis (see Section \ref{results:kmeans}) showed us that it is likely that the data may belong to a different number of clusters. Hierarchical clustering is generally used when the number of clusters is not known in advance and a distance threshold (cut-off threshold) can be used on a dendrogram to obtain a fixed number of clusters. For simplicity and consistency, we tune a single hyperparameter. We specify a range of values for \texttt{n\_clusters} for which the algorithm automatically decides the cut-off threshold to obtain the number of clusters.
\par
\begin{figure}[!htb]
    \centering
    \includegraphics[width=0.8\textwidth]{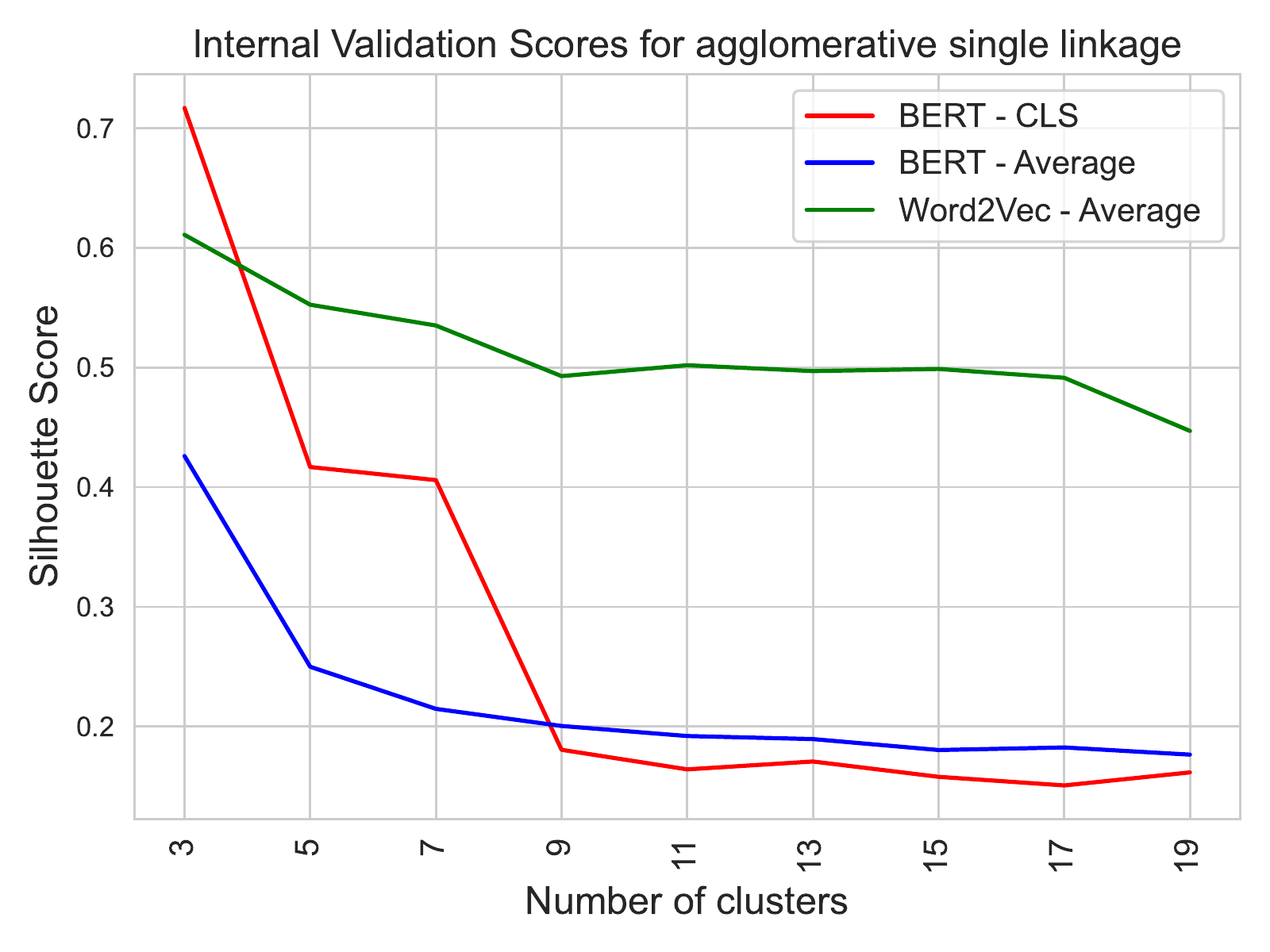}
    \caption{Internal Validation scores for Single Linkage Agglomerative Hierarchical Clustering}
    \label{fig:internal_validation_agglomerative}
\end{figure}

Similar to the analysis in Section \ref{results:kmeans}, we use the range of 3 - 19 for the hyperparameter \texttt{n\_clusters}. We show the internal validation results in Figure \ref{fig:internal_validation_agglomerative}. For BERT - CLS, the single linkage agglomerative clustering algorithm outperforms other embeddings when the number of clusters is 3. However, this effect is transient and the silhouette score drops quickly when the number of clusters is 5. We also observe a similar effect when the model is trained on the BERT - Average embeddings. On the other, when using Word2Vec - Average embeddings, the silhouette score of the model is greater than those obtained when the model is trained on BERT - CLS and BERT - Average embeddings when the number of clusters is greater than 3. Such a result indicates that the Word2Vec embeddings may be distributed in the space in a manner that enables the clustering algorithm to find clusters that have good cohesion and are well relatively well separated from other clusters in the space. However, since the single-linkage hierarchical clustering only considers the closest points between two clusters before merging them, it may cause the clusters to be connected in a chain-like manner. This may also result in the silhouette scores becoming relatively stable even if the number of clusters increases. \par

\begin{figure}[!htb]
    \centering
    \includegraphics[width=0.8\textwidth]{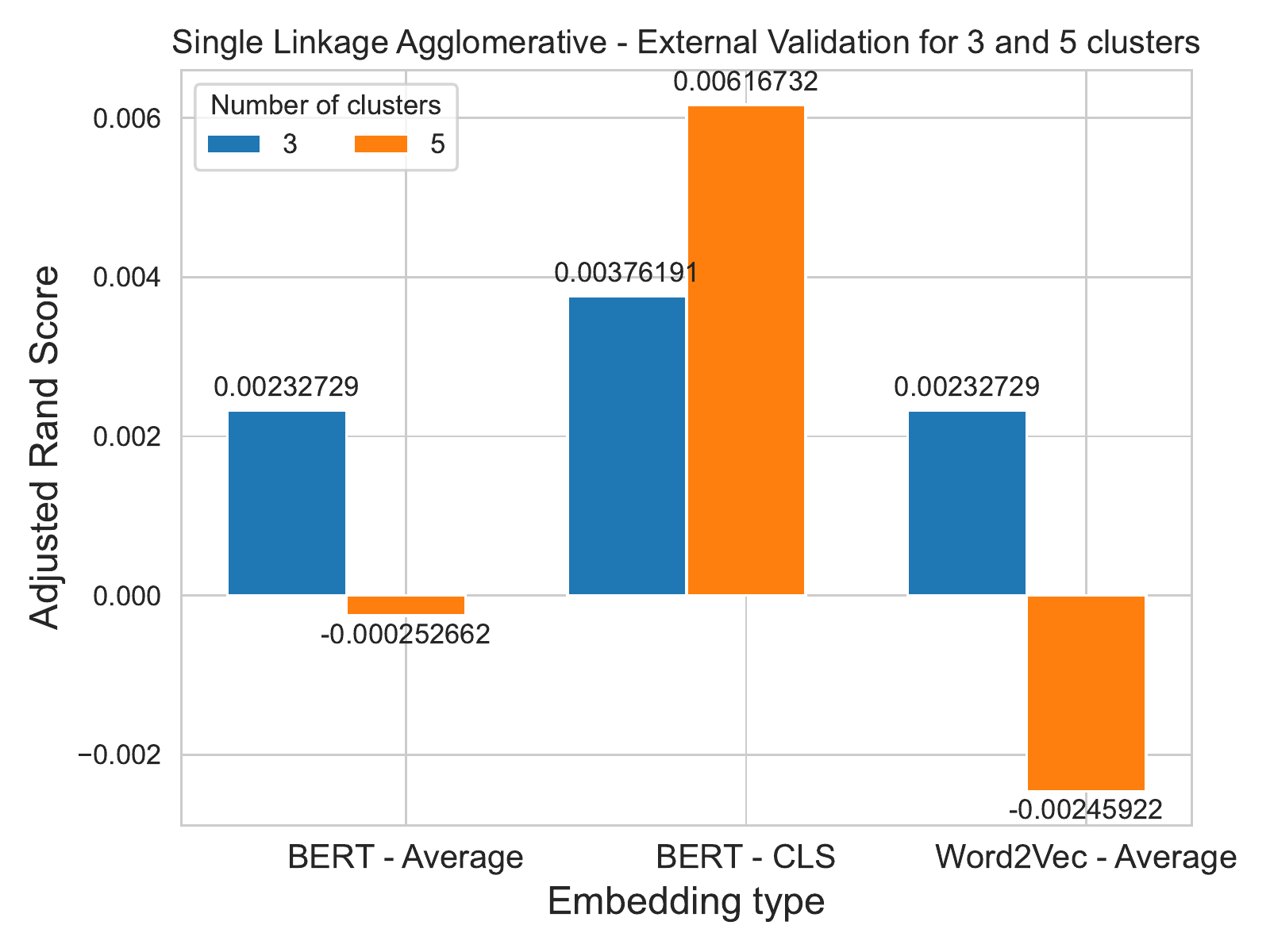}
    \caption{External validation scores for Single Linkage Agglomerative Hierarchical Clustering}
    \label{fig:external_validation_agglomerative}
\end{figure}

When comparing the silhouette scores for the single linkage agglomerative hierarchical clustering to those observed during the KMeans analysis, the silhouette scores for the single linkage agglomerative clustering are higher than those identified by the KMeans algorithm. Although, this may indicate that the single linkage agglomerative clustering performs better than KMeans, the external validation results in Figure \ref{fig:external_validation_agglomerative} indicated poor clustering quality as the adjusted rand scores were close to a value of zero. \par

\begin{figure}[!htb]
    \centering
    \includegraphics[width=0.8\textwidth]{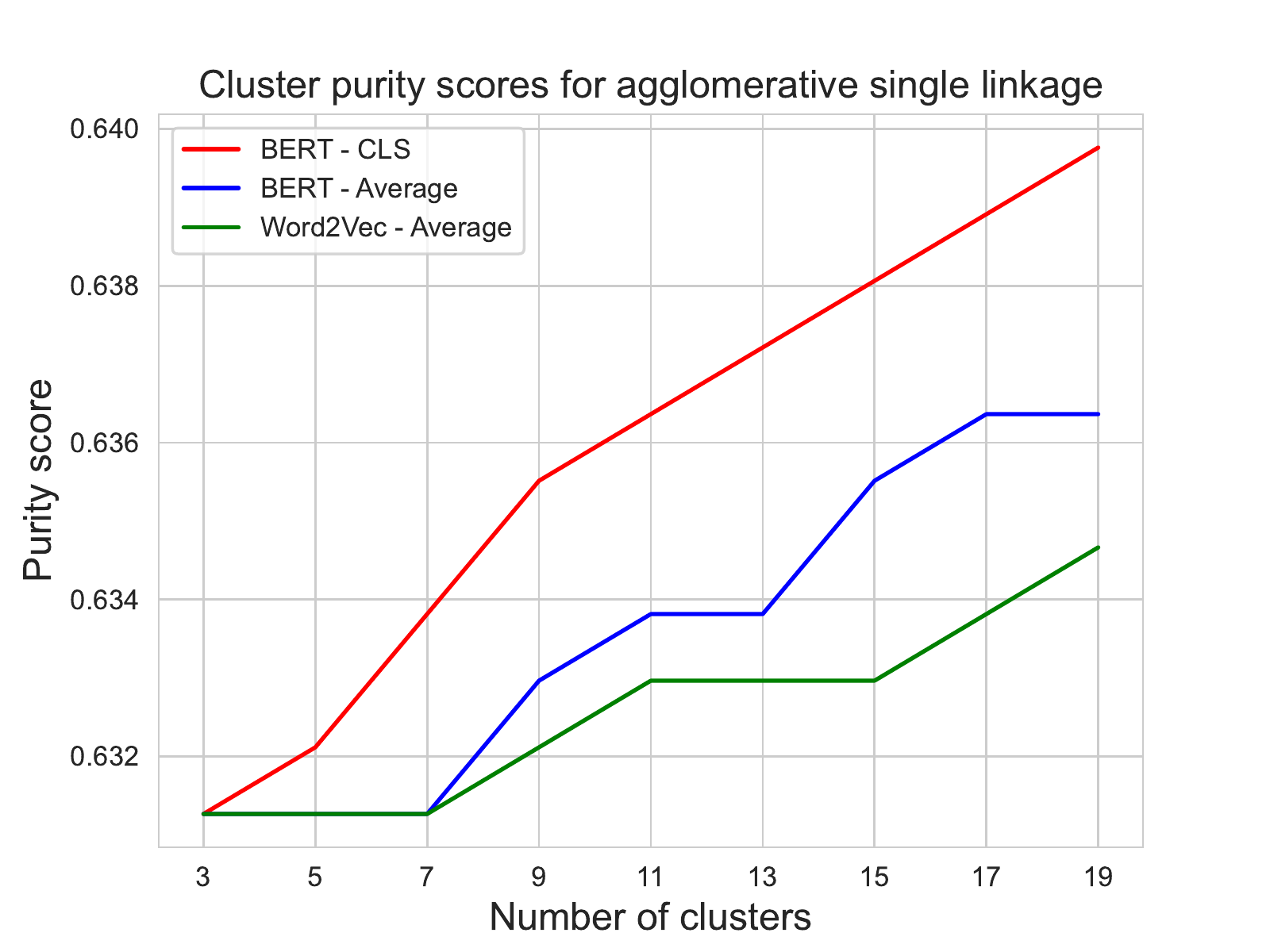}
    \caption{Cluster purity scores for DBSCAN. For each embedding type, the cluster purity increases with increase in the number of clusters.}
    \label{fig:purity_agglomerative}
\end{figure}

The cluster purity scores observed in Figure \ref{fig:purity_agglomerative} also suggested the poor clustering quality of the single linkage agglomerative clustering algorithm. Similar to the KMeans results, the purity scores were in the range of between 0.60 to 0.67 further providing evidence that the single linkage agglomerative hierarchical clustering algorithm is not better at finding clusters in the underlying data space.

\subsection{DBSCAN}
\label{results:dbscan}
\begin{figure}[!htb]
    \centering
    \includegraphics[width=0.8\textwidth]{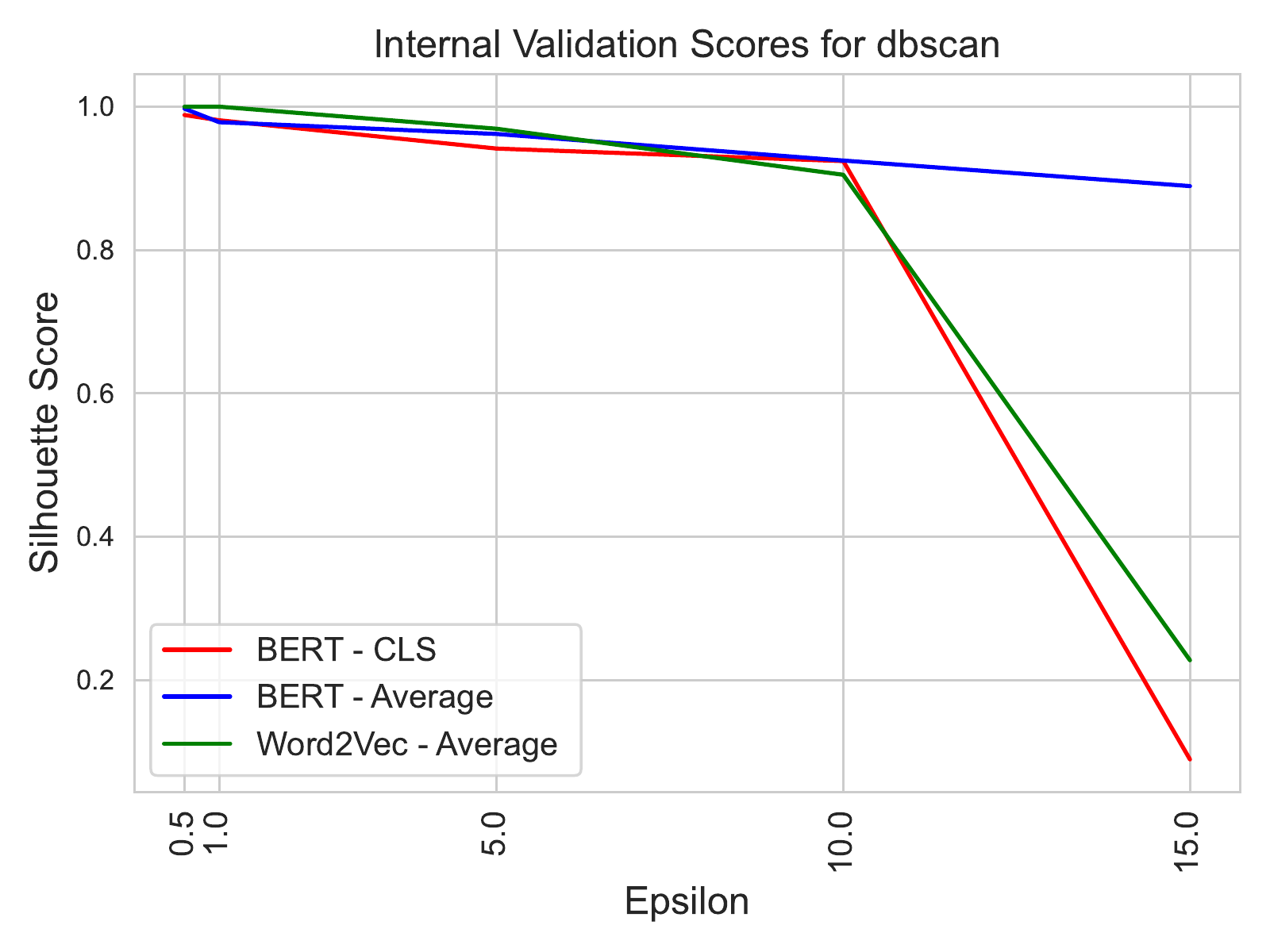}
    \caption{Internal validation scores for various epsilon values for the DBSCAN algorithm. Values of epsilon chosen were [0.5, 1.0, 5.0, 10.0, 15.0].}
    \label{fig:internal_validation_dbscan}
\end{figure}

Besides exploring KMeans and single linkage hierarchical clustering, we wanted to apply the density based algorithms 
To evaluate the density density based algorithms, we only use the silhouette score as the internal validation metric.
As reviewed previously, we only tune the value of the epsilon hyperparameter and keep the value of the \texttt{min\_samples} hyperparameter fixed to 5. For the DBSCAN algorithm, we show the results in Figure \ref{fig:internal_validation_dbscan}. We observe silhouette scores close to 1.0 for lower values of epsilon suggesting that DBSCAN can find well defined clusters and performs better than KMeans and single linkage agglomerative clustering. However, the number of clusters identified by DBSCAN is not equal to those identified by KMeans and single linkage agglomerative clustering. We saw in Sections \ref{results:kmeans} and \ref{results:agglomerative} that a lower number of clusters resulted in higher silhouette scores; around 3 clusters. On the other hand, for DBSCAN, we observed more clusters for higher silhouette scores. For epsilon=0.5, there were 13 clusters for the BERT - Average embeddings, 16 for BERT - CLS embeddings, and 22 for Word2Vec - Average embeddings. These results suggest that it is not always the case that model performance is consistent across different algorithms when applied to the same data. \par

There was also an interesting effect in terms of the change in the silhouette score when the value of epsilon was set to 15.0. The silhouette scores dropped drastically for BERT - CLS, and Word2Vec - Average. However, the silhouette score for BERT - Average did not drop drastically. This may be due to the underlying distribution of the BERT - Average embeddings in the data space that enables the algorithm to identify clusters that are more well defined than those of BERT - CLS and Word2Vec - Average embeddings. Therefore, one may further investigate the possibility of using BERT - Average embeddings when using DBSCAN to cluster textual review data.\par

Such results may suggest that indeed more clusters exist in the data. But one must be cautious before drawing conclusions as the number of points in each identified cluster may be low in number. It must be noted that while training the DBSCAN algorithm, we fixed the value of \texttt{min\_samples} to 5, which may have affected the results. Moreover, we also observed that when $\epsilon$ was set to 0.5, many points are identified as noise / outliers. 
We show the number of noise points for $\epsilon=0.5$, and $\epsilon=15.0$ in Table \ref{tab:noise_points_for_dbscan}. Therefore, each cluster contains very few points. 

\begin{table}[!htb]
    \centering
    \begin{tabular}{c|c|c}
     Embedding type & $\epsilon=0.5$& $\epsilon=15.0$ \\ \hline
     BERT - CLS & 1056 & 574\\
     BERT - Average & 1099 & 931\\
     Word2Vec - Average & 973 & 376\\
    \end{tabular}
    \caption{Number of noise points for DBSCAN for $\epsilon$ values 0.5 and 15.0. }
    \label{tab:noise_points_for_dbscan}
\end{table}

The internal validations scores and the number of outliers for each embedding type suggest that even though the silhouette scores may appear to be high, the presence of outliers does not necessarily make the DBSCAN algorithm \textit{better} than the KMeans or the single linkage agglomerative hierarchical algorithms. One of the possible explanations for obtaining such results could be related to the distribution of the data points in the embedding space and we discuss this point in more detail in the limitations section (Section \ref{sec:limitations}).

\begin{figure}[!htb]
    \centering
    \includegraphics[width=0.8\textwidth]{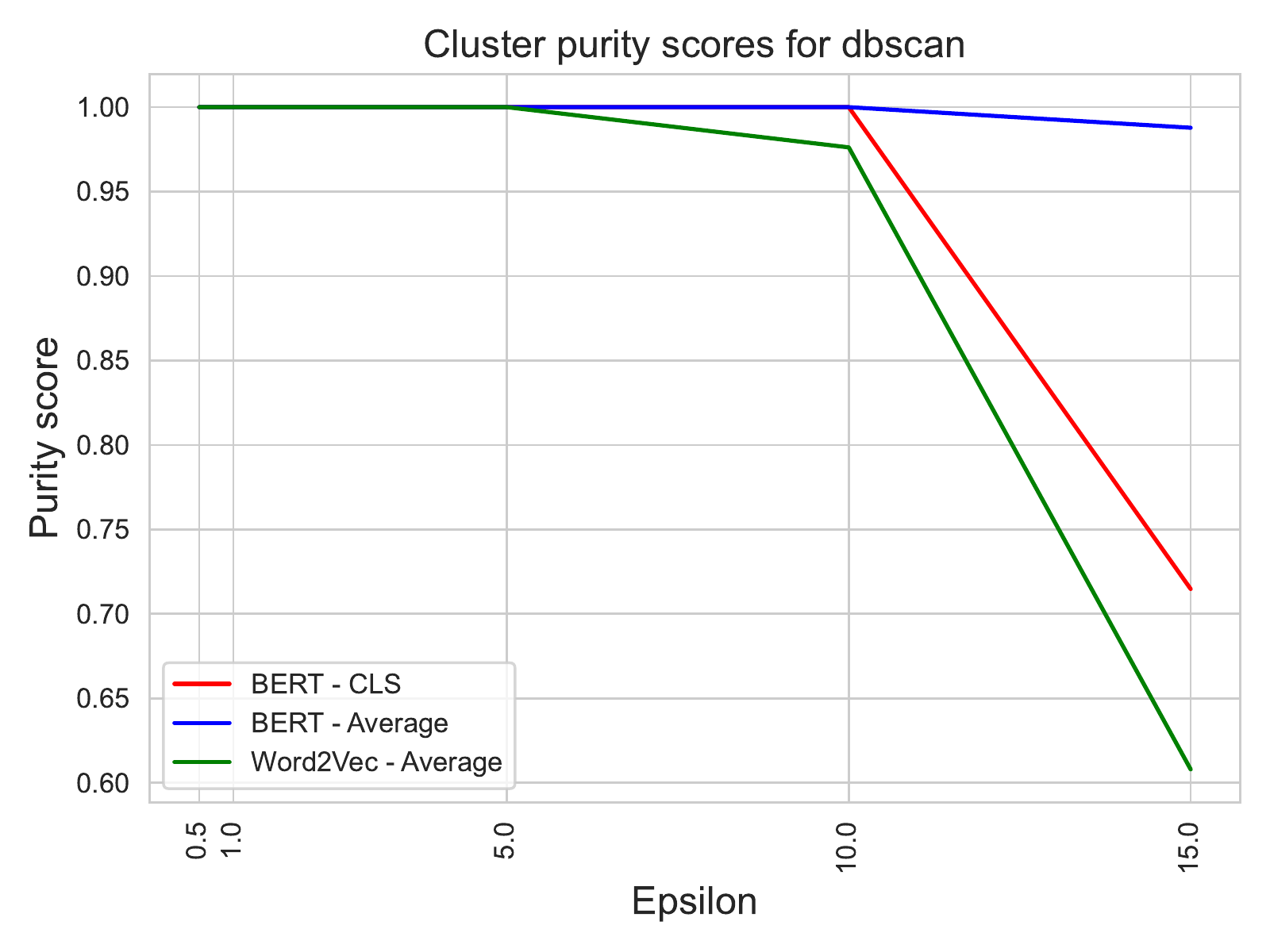}
    \caption{Cluster purity scores for the DBSCAN algorithm.}
    \label{fig:dbscan_purity}
\end{figure}

The cluster purity results shown in Figure \ref{fig:dbscan_purity} provide further support to the idea that for lower values of epsilon (and keeping the value of \texttt{min\_samples} fixed to a value of 5), smaller clusters are generated and most of the data points are labelled as noise points; as cluster purity is calculated only on the identified clusters, the value is high. Furthermore, when $\epsilon=15.0$, the number of noise points decreases indicating that more points are assigned to a cluster, but a decrement in the silhouette score observed in Figure \ref{fig:internal_validation_dbscan} suggests a poor clustering performance.

All in all, the results indicate that the DBSCAN algorithm's performance is highly pertinent to the distribution of the embeddings, the value of $\epsilon$, and the value of \texttt{min\_samples}. Although the silhouette and the cluster purity scores can be used to evaluate the performance of DBSCAN, one must also consider calculating the number of noise points as the relative proportion of noise points identified by the algorithm can affect the final interpretation of the results.

\subsection{HDBSCAN}
\label{results:hdbscan}
\begin{figure}[!htb]
    \centering
    \includegraphics[width=0.8\textwidth]{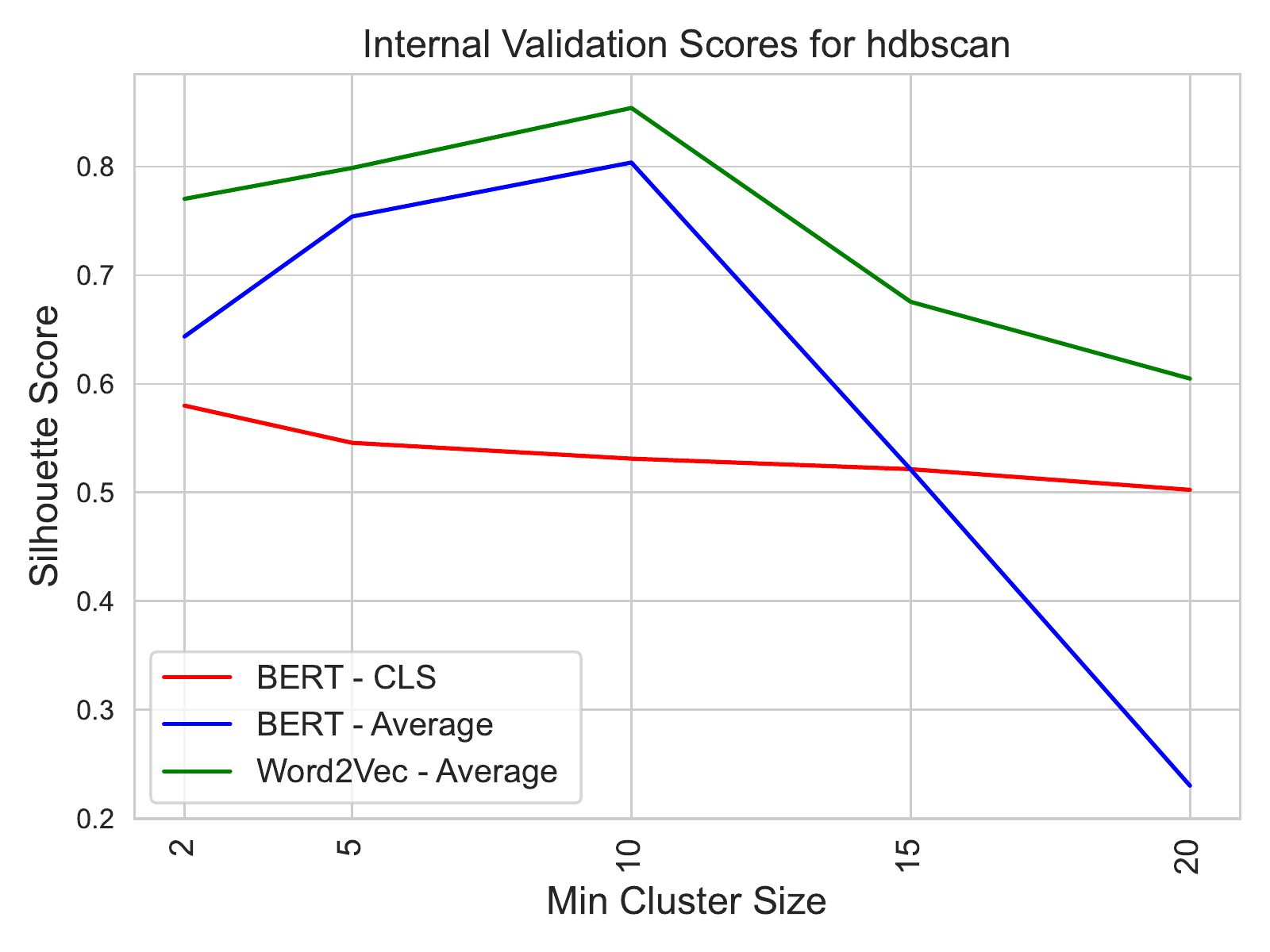}
    \caption{Internal validation scores for the HDBSCAN algorithm. Values chosen for the hyperparameter \texttt{min\_cluster\_size} were [2, 5, 10, 15, 20].}
    \label{fig:internal_validation_hdbscan}
\end{figure}

One of the shortcomings of DBSCAN is its inability to identify clusters of varying densities. As reviewed in Section \ref{sec:embeddings}, the distribution of embeddings in space can vary in density. Consequently, we also analyze the performance of HDBSCAN for clustering the various embedding types. We show the internal validation results for HDBSCAN in Figure \ref{fig:internal_validation_hdbscan}.
\par

For simplicity, we only tune the hyperparameter \texttt{min\_cluster\_size} of the HDBSCAN model, which controls the minimum size of the clusters below which all points are considered to be noise points, and keep all the other hyperparameters to their respective default values (see Section \ref{sec:clustering_algorithms}). We observe that when the model is trained on Word2Vec - Average and BERT - Average, the silhouette score is maximum when the \texttt{min\_cluster\_size} is 10. In addition, the silhouette score for the model trained on BERT - Average embeddings dropped sharply when the value of \texttt{min\_cluster\_size} was above 10. However, such an effect is not observed when the model is trained on the BERT - CLS embeddings. The silhouette scores for BERT - CLS were relatively stable for all values of \texttt{min\_cluster\_size}. These results suggest that when using algorithms such as HDBSCAN, average embeddings such as BERT - Average or Word2Vec - Average might be better suited for clustering text embeddings, instead of using BERT - CLS embeddings. \par
\begin{figure}[!htb]
    \centering
    \includegraphics[width=0.8\textwidth]{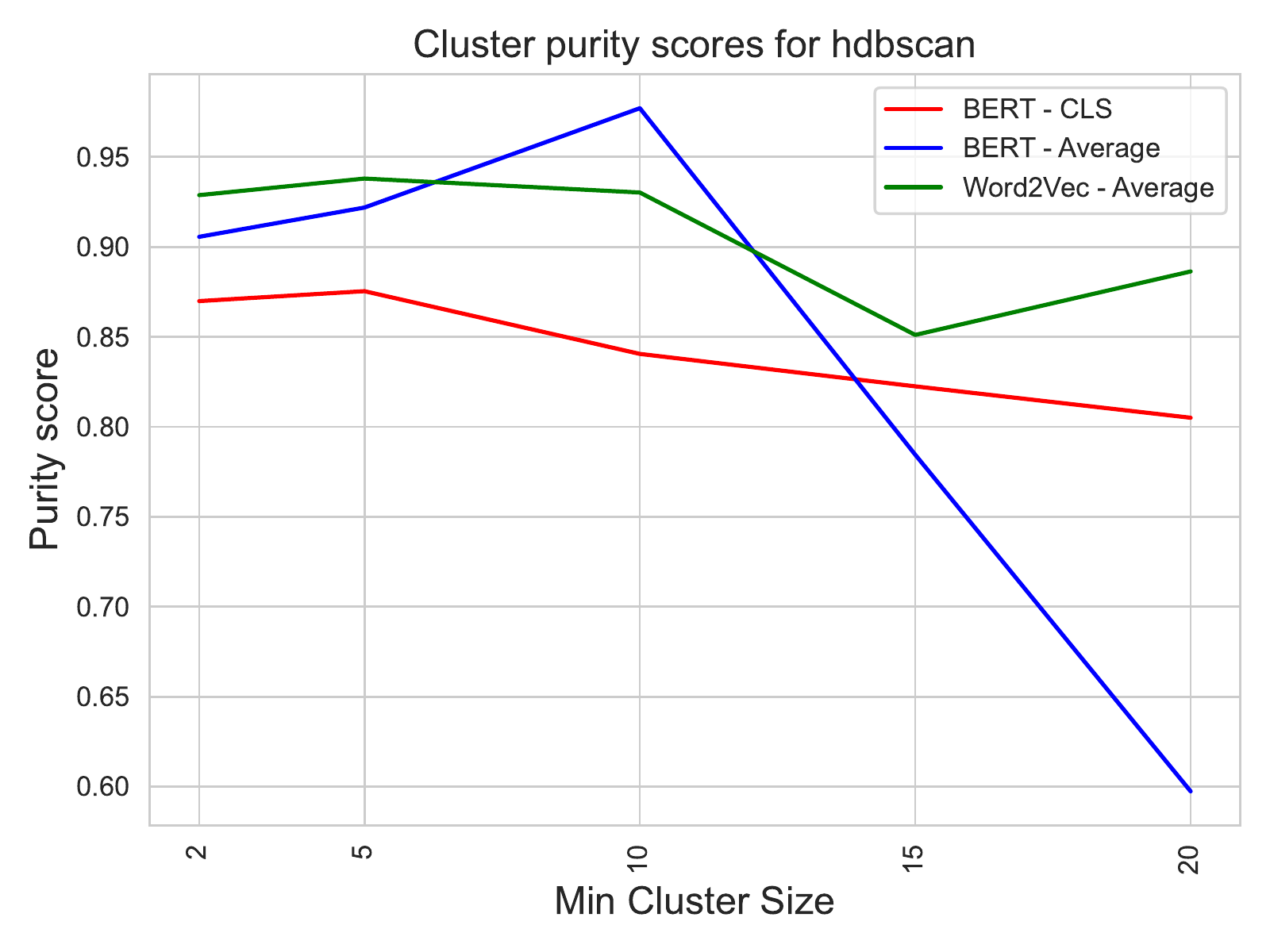}
    \caption{Cluster purity scores for the HDBSCAN algorithm. Values chosen for the hyperparameter \texttt{min\_cluster\_size} were [2, 5, 10, 15, 20].}
    \label{fig:purity_hdbscan}
\end{figure}

Do the silhouette scores reflect the clustering quality? We observe in Figure \ref{fig:purity_hdbscan} that when using the BERT - Average and Word2Vec - Average embeddings, HDBSCAN clusters have higher purity when using BERT - CLS embeddings. Generally, the clustering purity score is above 0.8 for all embeddings (except after when the \texttt{min\_cluster\_size} value is more than 10 in case of BERT - Average) suggesting that the clusters are of generally good quality. However, interesting results were observed when comparing the number of clusters identified by DBSCAN and HDBSCAN. The number of clusters identified by HDBSCAN was lower than those identified by DBSCAN. The number of clusters identified by HDBSCAN for each embedding is given below (when the \texttt{min\_cluster\_size} is 10).
\begin{itemize}
    \item BERT - CLS = 9
    \item BERT - Average = 7
    \item Word2Vec - Average = 7
\end{itemize}

Even for a lower number of clusters, the silhouette scores and the cluster purity scores are relatively high indicating that HDBSCAN may be more robust in clustering text embeddings. However, one must be cautious before drawing conclusions as the number of noise points identified by HDBSCAN may inflate the results. In fact, we observed high number of noise points identifed by HDBSCAN when \texttt{min\_cluster\_size} = 10; the number of outliers identified by HDBSCAN were 945 for the BERT - CLS embedding, 1046 for the BERT - Average embedding, and 1048 for the Word2Vec - Average embedding.
The number of outliers identified by HDBSCAN for the best performing hyperparameter are comparable to the number of outliers identified by HDBSCAN. In other words, both the density based algorithms do not appear to be significantly outperform the other.
\par
All things considered, when comparing all the above algorithms, DBSCAN and HDBSCAN appear to outperform KMeans and single linkage agglomerative hierarchical clustering algorithm. However, one must calculate the number of outliers identified by the density based algorithms as it may affect the final conclusion of the comparative analysis.

\section{Limitations and Future Work}
\label{sec:limitations}
Our work has potential limitations. We used a fairly restricted dataset that primarily has examples of consumer electronics from one brand. Therefore, minor variations in the performance of the clustering algorithms are plausible if a larger and more diverse dataset is used. An extension of this work may include a comprehensive comparison across various types of datasets (containing other types of items, multiple languages etc). \par
We did not factor in the demographics of the reviews. Reviews for the same product can exist in multiple languages and existing language models may represent a single review differently in different languages, which may lead to variance in the performance of the clustering algorithms. In other words, progress in optimal representations of text is a bottleneck when evaluating embeddings for clustering algorithms. This may be considered as a motivation for fostering more research in aligning text representations across multiple languages.

Users assign a star rating for multiple reasons and a product may be assigned a one star review for several reasons (bad packaging, dislike of a specific feature etc.). In this project, we treat all reviews for a given rating equally regardless of the underlying text / reason of the dataset in the review for that specific rating. However, customers assign a specific rating value for numerous reasons. For example, a customer may assign a product a rating of 5 because they liked the battery capacity. Another customer may have assigned a 5 star rating because the product was compact in size. This may be one of the reasons that our external validation scores were close to zero, as the assignment of ratings on a scale of 1-5 may be insufficient / too small of a range. Moreover, the number of clusters detected by the density based algorithms in Sections \ref{results:dbscan} and \ref{results:hdbscan} support the presence of more than 5 clusters in the underlying review dataset. Moreover, as many points were identified as noise points by the DBSCAN algorithm, it is possible that more underlying clusters may exist in the data space. Further work is necessary to evaluate clustering algorithms on clustering review data with the consideration of the number of clusters being more than the number of predefined labels. \par

This work uses three different evaluation metrics as interpreting clustering quality based on only one metric may not be accurate, especially when dealing with complex or high-dimensional data. However, there are advantages and limitations to all the metrics. Therefore, it is also possible that in the future more sophisticated metrics specifically designed for text embeddings, or a combination of existing metrics can be used to evaluate clustering results.
\par
While training the clustering algorithms, a limited number of hyperparameters were tuned for simplicity. More sophisticated hyperparameter tuning will help us gain a deeper understanding of the performance of the clustering algorithms given high-dimensional text embeddings. Future work may explore the ideas in this paper by tuning multiple hyperparameters for various algorithms. \par
To represent the text reviews, we used embeddings that were obtained from pretrained models (Word2Vec pretrained on the google news dataset, and the BERT model pretrained on the BookCorpus and English Wikipedia dataset), which may not be the best representative of the data in the review dataset. In other words, it might have been possible that fine-tuning the pretrained language models on the review dataset might have resulted in embeddings that are better suited for clustering. In the future, researchers may look at fine-tuning language models on multiple diverse datasets to obtain embeddings that can more accurately represent textual reviews than pretrained models. \par
Finally, it is useful to note that the choice of algorithm is crucial for clustering text data and this choice requires careful planning. For example, our results showed that DBSCAN outperforms the KMeans and single linkage hierarchical algorithm, but there were also a high number of noise points identified by DBSCAN. As the $\epsilon$ value increases, the number of noise points decreases, but the silhouette score and the cluster purity score also decrease (for all embeddings except BERT - Average). Such results may help require to motivate research into developing clustering algorithms specifically designed to cluster embeddings. Although some limitations exist, our work also presents numerous ideas that may be investigated in the future to advance research in the domain of text clustering.

\section{Conclusion}
\label{sec:conclusion}
In this work, we investigated the performance of various types of algorithms for clustering text embeddings and how choosing different types of embeddings impacts the clustering performance. We used a dataset containing product reviews listed on Amazon.com where each review was assigned a rating from 1 - 5. We represented the reviews using Word2Vec and BERT embeddings. We used the KMeans, Single Linkage Agglomerative Hierarchical clustering, DBSCAN, and HDBSCAN. We evaluated each algorithm using various measures such as the silhouette score, adjusted rand index score, and cluster purity. We observed that density based algorithms generally seem to outperform the KMeans and the single linkage hierarchical algorithms. In addition, the results of the density based algorithms suggested that the number of underlying clusters in the dataset may be more than the number of labels in the dataset (3 or 5 for the dataset). However, our results also highlight the challenges in clustering high-dimensional text data, such as the varied distribution of embeddings in space for the same rating value. One must also be careful about drawing conclusions when comparing clustering algorithms as different hyperparameter settings may have an effect on the results. For example, tuning the $\epsilon$ value of the DBSCAN algorithm and \texttt{min\_cluster\_size} for HDBSCAN (Section \ref{results:dbscan}) resulted in a different number of noise points being identified by the algorithms. This affects the clustering algorithm's performance. \par

All in all, this work provides a proof-of-concept where we carry out a comparative analysis of different textual embeddings' impact on the performance of clustering algorithms. We hope that this work opens up new avenues for research in the domain of text clustering and developing new embeddings to adequately represent text data.


\printbibliography

\end{document}